\documentclass{article}

\usepackage{microtype}
\usepackage{graphicx}
\usepackage{amssymb}
\usepackage{tikz-cd}
\usepackage[all]{xy}
\usepackage{amsmath}
\usepackage{subfig}
\usepackage{bm}
\usepackage{booktabs} % for professional tables
\usepackage[utf8]{inputenc} % allow utf-8 input
\usepackage[T1]{fontenc}    % use 8-bit T1 fonts
\usepackage{hyperref}       % hyperlinks
\usepackage{url}            % simple URL typesetting
\usepackage{booktabs}       % professional-quality tables
\usepackage{amsfonts}       % blackboard math symbols
\usepackage{nicefrac}       % compact symbols for 1/2, etc.
\usepackage{microtype}      % microtypography
\graphicspath{ {images/} }

\DeclareMathOperator{\ima}{im}
\renewcommand\vec{\bm}

\newsavebox{\measurebox}

    \usepackage[preprint, nonatbib]{neurips_2019}

\title{Characterizing the Shape of Activation Space in Deep Neural Networks}

\author{%
  Thomas Gebhart \\
  Department of Computer Science\\
  University of Minnesota\\
  Minneapolis, MN 55455 \\
  \texttt{gebhart@umn.edu} \\
   \And
  Paul Schrater \\
  Department of Computer Science\\
  University of Minnesota\\
  Minneapolis, MN 55455 \\
   \AND
   Alan Hylton \\
   NASA Glenn Research Center \\
   Cleveland, OH 44135 \\
}

\begin{document}

\maketitle

\begin{abstract}
    % Problem
    % gap in current knowledge
   The representations learned by deep neural networks are difficult to interpret in part due to their large parameter space and the complexities introduced by their multi-layer structure. We introduce a method for computing persistent homology over the graphical activation structure of neural networks, which provides access to the task-relevant substructures activated throughout the network for a given input. This topological perspective provides unique insights into the distributed representations encoded by neural networks in terms of the shape of their activation structures. We demonstrate the value of this approach by showing an alternative explanation for the existence of adversarial examples. By studying the topology of network activations across multiple architectures and datasets, we find that adversarial perturbations do not add activations that target the semantic structure of the adversarial class as previously hypothesized. Rather, adversarial examples are explainable as alterations to the dominant activation structures induced by the original image, suggesting the class representations learned by deep networks are problematically sparse on the input space.
    % take home message
    
    %In feedforward networks, information (in the form of activations) must survive to the final layer to contribute to classification.
\end{abstract}

\section{Introduction}
%% HERE IS A GOOD RECENT ICLR PAPER: https://openreview.net/forum?id=HylTBhA5tQ
% The intro to this paper is especially relevant
% See https://arxiv.org/pdf/1810.01185.pdf for a detailed explanation of the hypothesis space.

Neural networks are a class of learning model often characterized by the layer-wise composition of nonlinear functions parameterized by neurons which pass activation values through weighted connections between layers. This structure allows these models a high degree of expressivity in terms of transformations of the input space or task-specific representations learnable by the model, resulting in state-of-the-art performance in complex domains where relevant features are difficult to construct \emph{a priori}. 

The expressivity of these models comes at the cost of interpretability. With tunable parameters numbering in the millions for even modestly sized neural networks \cite{krizhevsky2014one}, it is difficult to probe the representations constructed by these models, and extracting reasoning for a specific network decision given an input harder still. Lacking this interpretability, neural networks elicit a range of peculiar behaviors, most striking of which are their susceptibility to adversarial examples.

We present a method for computing persistent homology over the activation graph of a neural networks. The output is a graded set of subgraphs which we find to be intimately related to the task-specific semantics learned by the network. We discuss the benefits of such a decomposition prove its use by studying the peculiarities of learned representations within neural networks and the extent to which adversarial examples exploit these representations within the network.

\subsection{Related Work}

Prior work investigating neural network representations generally takes the form of representation similarity analysis. In \cite{li2015convergent}, the authors investigate how to align neural network features based on their activations according to layer-wise bipartite and spectral clustering matching between activations, and note that representation codes are primarily distributed across neurons and the basis vectors spanned by individual units are not necessarily unique across networks, even if architecture is held constant. The understanding that representations are distributed is an important motivating factor for the methods presented in this paper. Similar approaches to understanding neural network activations through methods related to representation similarity analysis and cannonical correlation analysis are numerous \cite{raghu2017svcca,morcos2018insights,wang2018towards, khaligh2014deep, saxe2019mathematical}). 

% Adversarial examples have become an important issue in machine learning research due to the potential risks to security and robustness of machine learning systems, spawning efforts to design optimization procedures with robustness guarantees \cite{Kolter2018ProvableDA,JD2018ImprovingTG}, and theoretical work on the robustness and the cost landscape \cite{Wang2018TowardsRD}. Despite the intense research, there is substantial controversy about why adversarial attacks work, with hypotheses ranging from optimization, to generic properties of deep networks (e.g. the linearity hypothesis, to intrinsic characterizations of the data manifold \cite{Serban2018AdversarialE}). Recently, Gilmer et al \cite{Gilmer2018AdversarialS} have shown that adversarial examples are close in input space to correctly classified points while \cite{goodfellow2014explaining} note that the direction one moves to find an adversarial example is important, suggesting a complex manifold geometry for non-adversarial examples in the input space that might result from approximating the input distribution by a limited set of training examples.

The application of tools from topology, especially persistent homology, to analyze neural networks is relatively new. Recently, persistent homology has been used to analyze the parameter space of neural networks, especially during the training process \cite{guss2018characterizing, gabella2019topology}. Most similar to the current work is \cite{rieck2018neural} wherein the authors take a similar approach to constructing the network graph and associated filtration, but analyze the parameters of the network during training as opposed to the induced graph. 

Recent publications providing progress towards a fundamental understanding of adversarial examples are numerous. Most notably, Katz et al. \cite{katz2017reluplex} and Weng et al. \cite{weng2018towards} have found that computing a provably secure region of the input space is approximately  computationally hard. Mahloujifar et al. \cite{mahloujifar2018curse} explain the prevalence of adversarial examples by making a connection to the ``concentration of measure'' in metric spaces. Recently, Zhange et al. \cite{zhang2019limitations} found that adversaries are more dense sufficiently far from the manifold of training data.

\section{Neural Network Topology}\label{Neural Network Topology}

By viewing a neural network as a weighted, undirected graph--a metric space--we can use persistent homology to investigate its invariant topological properties across all edge weight resolutions. We consider only forward architectures in this construction, but a similar approach may be used to investigate more complex architectures with, for example, recurrent structure. Background on persistent homology can be found in the Supplemental Materials as well as \cite{ghrist2008barcodes} and \cite{edelsbrunner2010computational}. 

There are two ways to construct a graph representation of a neural network. These are the \textit{static network} consisting of (possibly trained) weight matrices connecting each layer's nodes to the next layer's, and the \textit{induced network} consisting of this same structure augmented by the activation functions and an input applied to the neural network. In this paper, we are concerned with the latter induced graphical structure, but both structures may be used to provide insight into the function of neural networks.
\begin{figure}[t]
\begin{center}
\centerline{\includegraphics[height=6cm, keepaspectratio]{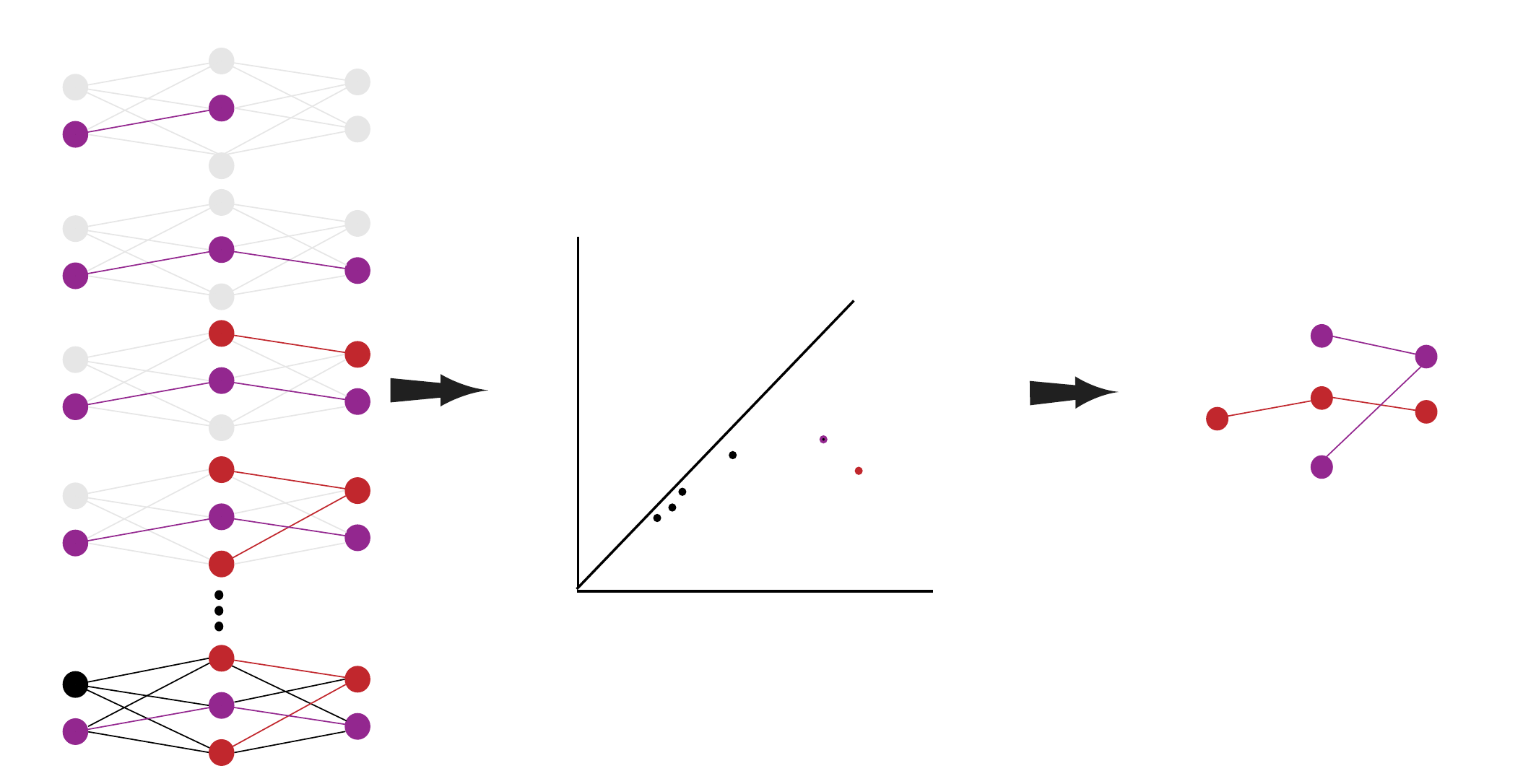}}
\caption{An illustration of neural network filtration and persistent subgraph reconstruction.}
\label{filtration}
\end{center}
\vspace{-0.5cm}
\end{figure}
Fix a feedforward neural network architecture with $L$ layers. Let $G^\mathcal{I} = (V,E,\phi)$ be the network's graphical representation \textit{induced} by input $\mathcal{I}$. Here $V$ is the set of nodes in the network including input, hidden, and output nodes, $E$ is the set of edges between nodes, and $\phi: E \rightarrow \mathbb{R}$ is a function assigning weights to edges. $G^{\mathcal{I}}$ is a multipartite graph where each node in a given layer may share an edge only with the previous and next layers' nodes. In other words, $V = V_0 \sqcup V_1 \sqcup \dots \sqcup V_{L-1}$ where $u \in V_k$, $v \in V_l$, and $(u,v) \in E$ only if $k= l-1$. Define $\bm{h}_l \in \mathbb{R}^{|V_l |}$ to be the activation values for each node in layer $l$. Define $h_u^l$ to be the $u$'th entry in vector $\bm{h}_l$. Let $\bm{W}_l$ be the weight matrix connecting the node set $V_l$ to $V_{l+1}$. This weight matrix is either the true weight matrix if layer $l$ is fully-connected, or if $l$ is a convolutional layer, the unrolled filter matrix described below. Let the entry in $\bm{W}_l$ corresponding to the $v$'th row and $u$'th column be denoted by $w_{u \rightarrow v}^l$. Again, we will drop this layer superscript unless otherwise necessary. With this information, we define the edge weighting for edge $(u,v) \in E$ via
\begin{equation*}
    \phi(u, v) = | w_{u \rightarrow v} h_u |
\end{equation*}
where $u \in V_l, v \in V_{l+1}$. For notational simplicity, we will drop the layer $l$ superscript when the layer is obvious. We use the absolute value of the hidden activation times the corresponding weight parameter to align the ordering of simplices in the filtration with the corresponding relationship to network semantics where (high activation or high suppression) are more important to classification decisions, whereas activation values close to $0$ are negligible in their semantic role. 

The filtration of $G^\mathcal{I}$ is defined via filtration on $\phi$. We have $N = |E|$ values of $\phi$ which can be ordered by $\geq$ such that $\max\limits_E \phi = \omega_0 \geq \omega_1 \dots \geq \omega_N = \min\limits_E \phi \geq 0$. The associated filtration on $G^\mathcal{I}$, $\emptyset \subset G^\mathcal{I}_0 \subset G^\mathcal{I}_1 \subset \dots \subset G^\mathcal{I}_N = G^\mathcal{I}$ is defined by adding the corresponding edge (1-simplex) and vertices (0-simplices) of each $\omega_i$ into the graph. In other words, $G^\mathcal{I}_0$ is the graph consisting of only the highest weight edge (the 1-simplex given by $\max\limits_E \phi$) along with the vertices (0-simplices) connected by that edge, $G^\mathcal{I}_1$ is the (potentially disconnected) graph consisting of the previously described edge and vertices along with the edge and vertices associated to $\omega_1$. 

The intuition for constructing the input-induced graph from a neural network is as follows. For each fully-connected layer, we connect each hidden node of layer $l$ to each hidden node in layer $l+1$ with an edge. The weight of this edge is the activation value of the hidden node multiplied by the corresponding element in weight matrix $\bm{W}_l$. The graphical construction for convolutional layers is similar but requires slightly more preprocessing. For each incoming channel, unroll the convolution operation into a (sparse) matrix multiplication operation. This multiplication operation operation then induces a graphical representation like in the fully-connected case. Because this operation is sparse, most of the edges have weight 0, indicating a stride of the filter whose preimage did not include that neuron. This sparsity can be easily filtered out as it does not affect the persistence calculation. Max pooling layers are constructed similarly, but the value passed from the node in the image of the filter is the maximum of the activations in the preimage.

\subsection{Persistent Subgraphs}\label{Persistent Subgraphs}

With a filtration on $G^\mathcal{I}$, we can compute its persistent homology. We say a homology class $\alpha$ is \textit{born} at $G^\mathcal{I}_i$ if it is not in the image of the map induced by the inclusion $G^\mathcal{I}_{i-1} \subset G^\mathcal{I}_i$. If $\alpha$ is born at $G^\mathcal{I}_i$, it \textit{dies} entering $G^\mathcal{I}_j$ if the image of the map induced by $G^\mathcal{I}_{i-1} \subset G^\mathcal{I}_{j-1}$ does not contain the image of $\alpha$ but the image of the map induced by $G^\mathcal{I}_{i-1} \subset G^\mathcal{I}_j$ does. We call $j - i$ the \textit{lifetime} or \textit{persistence} of the topological feature generated by $\alpha$. The filtration on $G^\mathcal{I}$ induces the sequence of homology groups $0 = H_p(\emptyset) \rightarrow H_p(G^\mathcal{I}_0) \rightarrow H_p(G^\mathcal{I}_1) \rightarrow \dots \rightarrow H_p(G^\mathcal{I}_N) = H_p(G^{\mathcal{I}})$. The lifetime of a feature born at level $i$ in the filtration and dying at level $j$ in the filtration is precisely the rank of the persistent homology group $H^{i,j}_p (G^\mathcal{I})$. For each dimension $p$, these persistent topological features can be represented as vectors in the half-plane via a persistence diagram (Figure \ref{fig:distances}) where more persistent features are located farther off the diagonal, while features that may be considered topological noise are associated to points near or along the diagonal. 

We are interested in zero-dimensional ($p=0$) topological information within this paper. Higher-dimensional topological features like holes and voids can be analyzed using a similar formulation, but the interpretation of these features with respect to neural network classification performance is less clear. By contrast, zero-dimensional topological features in this network space correspond to connected components, with the homology sequence describing how various components are created and merged with larger structures across various weight scales within the filtration. As shown below, these connected components capture important semantic information within the feature space of the neural network. 

Each $\alpha$ homology feature represents an equivalence class of some non-boundary cycle, so we can only pick a representative subgraph for a particular persistent connected component. However, because each of a typical neural network's (non-zero) weights are unique, we are nearly guaranteed that each equivalence class will contain only one element, namely the generator itself. In fact, a trained network that reliably contains non-trivial equivalence classes would only be optimal in a domain with high symmetry in input space. With stochastic weight updates and random initialization, the probability of encountering such a network is extremely low. In such a case, the persistent subgraphs corresponding to these generators can still be captured through the representative, but information on the number of connected components represented by this symmetric feature in input space is lost. 

With identification of the nodes and edges associated with each simplex in the filtration, we can reconstruct the subgraphs within the neural network that represent these persistent topological features. Let $A^{\mathcal{I}} = \{\alpha_1, \alpha_2, \dots \alpha_n\}$ be the set of $0$-cycle generators for $H_0(G^\mathcal{I})$. Each $\alpha_i$ is a simplicial complex that may be represented as some subgraph of $G^{\mathcal{I}}$. Equivalently, and under a slight abuse of notation, $\alpha_i = (V_i, E_i, \phi|_{E_i})$ where $V_i \subset V, E_i \subset E$.

%   \begin{figure}[h]
%  \centering
%   \includegraphics[width=0.23\textwidth]{CFFRelu/fashion/diagram}
%   \includegraphics[width=0.23\textwidth]{CFFRelu/fashion/adversary_diagram}
%   \includegraphics[width=0.23\textwidth]{CFFSigmoid/fashion/diagram}
%   \includegraphics[width=0.23\textwidth]{CFFSigmoid/fashion/adversary_diagram}
%  \caption{Persistence diagrams for networks CFF-Relu (top row) and CFF-Sigmoid (bottom row) persistent homology calculations induced by unaltered inputs (left column) and adversarial inputs (right column).}
%  \label{fig:Persistence Diagrams}
%  \end{figure}

%TODO 
% end each paragraph with a takeaway/actionable insight
\section{Topology of Neural Network Activations}
A critical reason for introducing the notion of induced network topology is that it allows one to view a network's response to a particular input at across multiple resolutions while retaining the nonlinearities of the network. Persistent homology gives access to a (po)set of graded subgraphs, related by an inclusion relationship, which make up the network's activation response to a particular input. As opposed to picking an activation threshold and viewing the network of activations above this threshold, persistence allows one to interrogate the network \textit{across} such thresholds while retaining information about how sub-networks at lower activation values are related to the structures that emerge at higher thresholds. This discrepancy is important for the analysis of deep neural networks wherein the range of activation values may be large but task-specific information may be distributed throughout a subgraph of lesser-valued activations whose combined effect leads to changes in the output class. The filtration of neural networks provided by persistent homology contains structure which is intimately related to the semantics of the classification task. The generators are linked to input features and the inclusion relationship over the generators induces a inclusion relationship over sets of features in the input space. The lifetime of these generators provides a measure of globality of subgraphs and, by extension, a measure of the globality of representations within the network. The specific structure of the generators in terms of their location in the larger network provides insight into the role of the generators in the output decision of the network. We discuss in this section how this persistence structure provides a unique and powerful viewpoint through which to analyze the representations learned by deep networks and find clarity in the potential causes of adversarial vulnerability in these models.  

% define a notion of inclusion in image space "input feature set" which follows the inclusion relationship of the generators

Persistent homology gives access to a set of graded activation subgraphs (the generators) distributed across the network architecture. Following this understanding, we can associate some meaning--via the semantics of the input space--to the generators $A^\mathcal{I}$. Each generator $\alpha_i$ corresponds to a feature within the input represented by the network. More precisely, recall from Section \ref{Persistent Subgraphs} that the filtration on $G^{\mathcal{I}}$ induces a sequence of (0-dimensional) homology groups $0 = H_0(\emptyset) \rightarrow H_0(G^\mathcal{I}_0) \rightarrow H_0(G^\mathcal{I}_1) \rightarrow \dots \rightarrow H_0(G^\mathcal{I}_N) = H_0(G^{\mathcal{I}})$. This induces the following relationship
\begin{equation*}
\xymatrix{
  0 \ar[r] & H_0(G^\mathcal{I}_0) \ar[d] \ar[r] & H_0(G^\mathcal{I}_1)  \ar[d] \ar[r] & \cdots \ar[r] & H_0(G^\mathcal{I}_{N-1}) \ar[d] \ar[r] & H_0(G^\mathcal{I}_{N}) \\
  0 \ar[r] & \mathcal{I}_0 \ar[r] & \mathcal{I}_1 \ar[r] & \cdots \ar[r] & \mathcal{I}_{N-1} \ar[r] & \mathcal{I}_N
}
\end{equation*}
for some filtration of the input $\emptyset \subset \mathcal{I}_0 \subset \mathcal{I}_1 \subset \dots \subset \mathcal{I}_N = \mathcal{I}$. The chain of homology groups (the inclusion of the generators) corresponds to an inclusion relationship for features in the input such that a lower-lifetime homology group dying via inclusion in a longer-lived group implies the subgraph representing the union of these two generators contains the input information from both. For example, a generator in the first layer of the network may represent the edge connecting a pixel within an input image to a weight within a convolutional filter. However, the inclusion of this edge into a generator with edges from other locations of the input as well as nodes in the next layer gives rise to a generator which captures the semantics within (spatially) disparate regions of the input along with information of how these features are combined by the network. This diagrammatic relationship mapping the filtration of the network to its semantic content provides a powerful abstraction for reasoning about the behavior of neural networks. Through proper characterization of the relationship between generators and their representation in input space, one can associate features of the input to their representational subgraphs within the network and begin reasoning about the robustness and versatility of these features in either network or input space. Crucially, the existence of a map between network structure and an input provides a way to compare representations \emph{across} networks by analyzing the filtration induced on the input. 

% But these subgraph structues represent functions \cite{farhoodi2019functions} computed by the network, so one could abstract this relationship to a sequence over the entire input

The generators also come equipped with a natural measure of globality through their lifetime. This structure can also be used for interpretation of network activations. Under this measure, an infinitely-lived generator represents a global descriptor of its constituent generators. For a well-trained network, we expect this generator to contain the correct class's neuron as a node in the subgraph it represents, as the activations flowing to this node should be the highest of all nodes in the final layer. In Section \ref{Experiments} we find that a network may have more than one infinitely-lived generator. One may interpret this behavior as the network \emph{suppressing} representations from other classes to arrive at a proper classification decision. However, it is unclear whether this behavior is always suppressive, in which case this behavior is indicative of the network carrying information deeper into layers than would otherwise be necessary for proper task performance given that input. The lifetime of the generators provides extra information for comparing the relative importance of two representations within a network and provides a measure to compute similarity between two persistence structures across scales.  % Determining the extent to which unnecessary information is propagated through the network would be an interesting avenue for future work.

% why does the lifetime matter, what are the nodes that it terminates in? lifetime gives locality versus globality. output semantics, input semantics, globality

To compute similarity in the space of persistent subgraphs, we use a hamming-like distance augmented by edge weights and lifetime of the generator within which that edge is located. That is, given the function $l(\alpha) = d(\alpha) - b(\alpha)$ mapping a generator $\alpha$ in the persistence diagram to its lifetime where $d$ and $b$ output positive real values, let the vector $\vec{o}({A^\mathcal{I}})_{A^{\mathcal{I}} \bigcup A^{\mathcal{J}}}$ be the one-hot-encoded vector for the set $A^{\mathcal{I}}$ of all edges across all graphs of $A^{\mathcal{I}}$ and $A^{\mathcal{J}}$ for two inputs $\mathcal{I}, \mathcal{J}$. This vector will have a $1$ for every edge in the generators of $A^{\mathcal{I}}$ and a $0$ everywhere the generators of $A^{\mathcal{J}}$ have an edge that is not in $A^{\mathcal{I}}$. For two of these vectors $\vec{o}({A^\mathcal{I}})_{A^{\mathcal{I}} \bigcup A^{\mathcal{J}}}$ and $\vec{o}({A^\mathcal{J}})_{A^{\mathcal{I}} \bigcup A^{\mathcal{J}}}$, we can calculate their hamming distance to arrive at a notion of the difference between two induced subgraph structures of the same network. However, this measure throws away the hierarchical information returned by persistent homology. We can instead compute the \emph{lifetime-weighted distance} between two induced subgraphs structures by replacing the $1$'s with the lifetime $l(\alpha)$ of the generator within which that edge is contained multiplied by the edge weight.

\section{Experiments}\label{Experiments}

% \begin{figure}[t]
% \centering
%         \raisebox{-\height}{\includegraphics[width=0.49\textwidth, keepaspectratio]{CFFSigmoid/fashion/distance/carliniwagnerl2/dist_to_adversary.png}}
%         \raisebox{-\height}{\includegraphics[width=0.49\textwidth, height=5cm, keepaspectratio]{CFFRelu/mnist/trajectory.png}}
% \caption{Left: $L_\infty$ distance between unaltered images and their adversarial counterparts for C\&W $L_2$ versus Wasserstein distance of their induced persistence diagrams within the CFF-Sigmoid network. Right: Trajectory of the persistence diagram as an MNIST image is transformed into an adversary.}
% \vspace{-0.5cm}
% \label{fig:distances}
% \end{figure}

\begin{figure}[t]
\centering
\sbox{\measurebox}{%
  \begin{minipage}[b]{.45\textwidth}
  \subfloat
    {\label{fig:figA}\includegraphics[width=\textwidth,height=5cm]{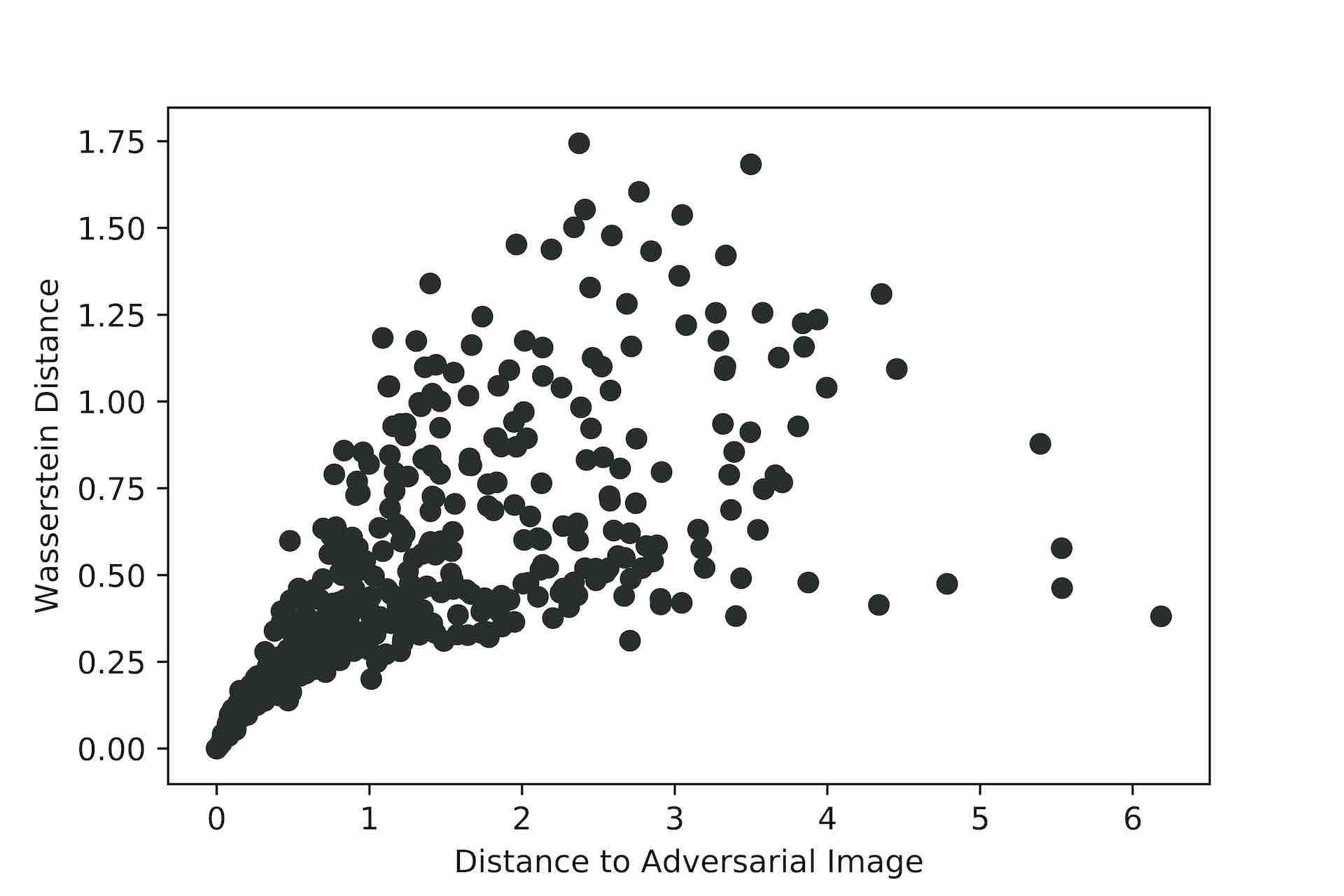}}
    \centering
  \end{minipage}}
\usebox{\measurebox}\qquad
\begin{minipage}[b][\ht\measurebox][s]{.39\textwidth}
\centering
\subfloat
  {\label{fig:figB}\includegraphics[width=0.49\textwidth,height=2.5cm, keepaspectratio]{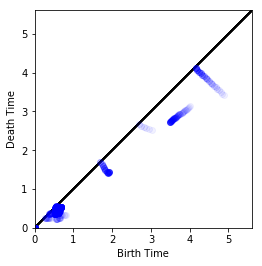}}
  {\label{fig:figB}\includegraphics[width=0.49\textwidth,height=2.5cm, keepaspectratio]{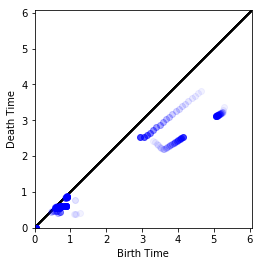}}
  
  {\label{fig:figC}\includegraphics[width=0.49\textwidth,height=2.5cm, keepaspectratio]{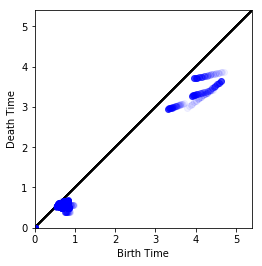}}
  {\label{fig:figC}\includegraphics[width=0.49\textwidth,height=2.5cm, keepaspectratio]{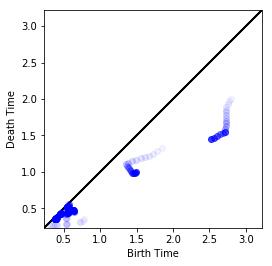}}
\end{minipage}
\caption{Left: $L_\infty$ distance between unaltered images and their adversarial counterparts for C\&W $L_2$ versus Wasserstein distance of their induced persistence diagrams within the CFF-Sigmoid network. Right: Trajectory of the persistence diagrams as MNIST images are transformed into adversaries.}
\vspace{-0.6cm}
\label{fig:distances}
\end{figure}

We implement the ideas provided in the previous sections using three neural network architectures of reasonable size and three datasets. These experiments highlight peculiarities in the representations learned by deep networks and suggest routes for improving their robustness and generalization capabilities. 

We trained three architectures, two on MNIST and Fashion MNIST, and one on the CIFAR10 dataset. The FashionMNIST and MNIST networks are both four layers, two convolutional followed by two fully-connected layers, but differ in their activation function. We refer to the network with ReLU activation as CCFF-Relu and the network with sigmoid activations as CCFF-Sigmoid. The CIFAR10 network is an AlexNet \cite{krizhevsky2014one} variant. See the supplementary materials for more information on network architectures. 

\subsection{Adversaries}\label{Adversaries}

For each network and dataset combination, we create a set of adversaries based off of the test set of each dataset. We create adversaries through two different methods. The first is the $L_2$ version of the Carlini Wagner adversaries described in \cite{carliniwagner} which we refer to as C\&W $L_2$. The other adversarial generation algorithm is the Projected Gradient Descent Attack as described in \cite{madry2017towards} with $\epsilon = 0.001$ and step size $0.01$ which we refer to as PGD.

%  \begin{figure}[h]
%   \includegraphics[width=0.49\textwidth]{images/CFFSigmoid/fashion/distance/carliniwagnerl2/dist_to_adversary.png}
%   \subfigure[]{
%   \includegraphics[width=0.49\textwidth]{images/CFFRelu/mnist/mnist_diff.png} \\
%   \includegraphics[width=0.49\textwidth]{images/CFFRelu/mnist/trajectory.png}
%   }
%   \vskip -0.1in
%  \caption{}
%  \label{fig:distances}
%  \end{figure}

\subsection{Subgraph Classification}\label{Subgraph Classification}

\begin{table}[t]
\caption{Classficiation accuracy for the Subgraph SVM classifier and the base accuracy of the neural network (Network Accuracy). Recovery Accuracy is the Subgraph SVM's accuracy in predicting the true class of an adversarial input to a network. Subgraph SVM accuracy is reported as the average of 10-fold cross-validations, and Recovery Accuracy is computed over 2000 adversarial examples. Top: MNIST. Bottom: Fashion MNIST.}
\label{tab:subgraph prediction}
\centering
\vskip 0.1in
\begin{tabular}{@{}llll@{}}
\toprule
Network     & Subgraph SVM Accuracy & Network Accuracy   & Recovery Accuracy \\ \midrule
CCFF-Relu    & 89.3\%                     & 97.6\%          &  70.3\% \\
CCFF-Sigmoid & 89.1\%                     & 88.8\%          & 83.4\% \\ \bottomrule

     &   &   & \\ 
CCFF-Relu    & 89.3\%                     & 90.0\%           & 80.3\% \\
CCFF-Sigmoid & 80.0\%                     & 80.2\%           & 73.3\% \\ \bottomrule
\end{tabular}
\end{table}

To show that homology reliably extracts activation subgraphs of the network that represent pathways of task-relevant information from input to output layer, we train a simple SVM using the lifetime-weighted distance as kernel over the persistent subgraph structure and show that a classifier based on this activation structure alone can achieve excellent classification performance (Table \ref{tab:subgraph prediction}). We also find that, given an adversarial input, this simple model can overwhelmingly recover the correct class of the input, implying the persistent structure of the network largely preserves the semantic content of the original image, despite misclassification by the network. We can see from Figure \ref{fig:distances} that even when restricted to only information within the persistence diagram (number of generators and their lifetimes), persistent homology still captures information about the input through the network activation graph. We see even for the more sophisticated C\&W $L_2$ attack a nearly linear relationship between $L_\infty$ distance in image space and Wasserstein distance between persistence diagrams for most adversaries, implying that persistent homology is able to capture information about the input space from the activation graph, even when the network itself misclassifies the input. These results imply that the underlying class information from the input is still recoverable from the network at a global scale, but local deviations in the network induced by adversarial perturbations cause misclassification.

\subsection{Nearest Neighbors}\label{Nearest Neighbors}
For a well-trained neural network $f$ with training data properly sampling the input  distribution, we expect the generators $\{A^\mathcal{I} \ | \ f(\mathcal{I}) = c\}$ induced by inputs predicted by the network to be of class $c$ to be related in their hierarchy derived from the inclusion structure in the generators. In other words, we expect the subgraph structure induced by inputs of the same class to be similar, sharing representation structures. We investigate this relationship by computing the nearest neighbors in input space and comparing them to the nearest neighbors terms of their induced persistent subgraphs using the lifetime-weighted distance over the subgraphs and their lifetimes. 

\begin{figure}
\centering
        \raisebox{-\height}{\includegraphics[width=0.32\textwidth]{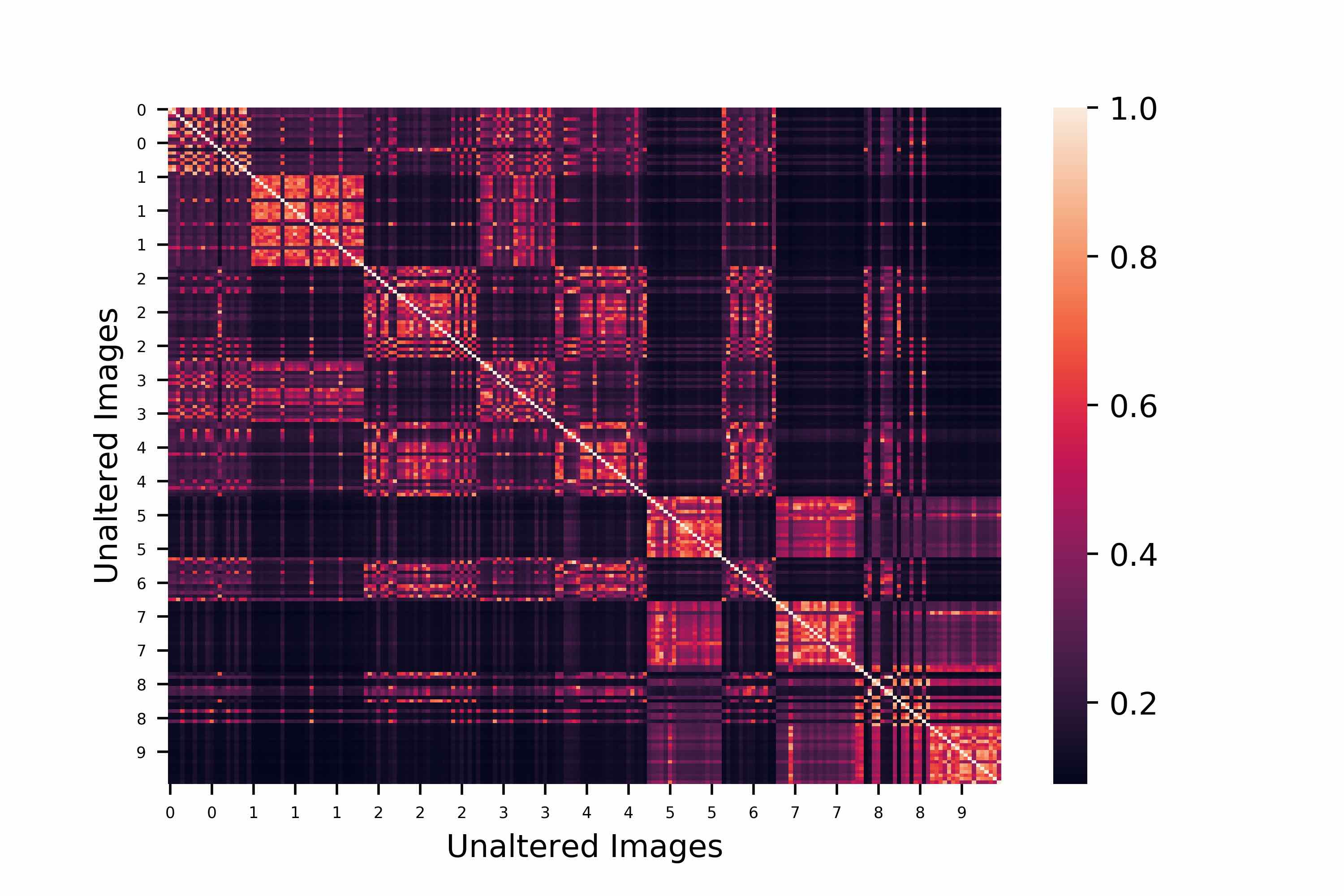}}
        \raisebox{-\height}{\includegraphics[width=0.32\textwidth]{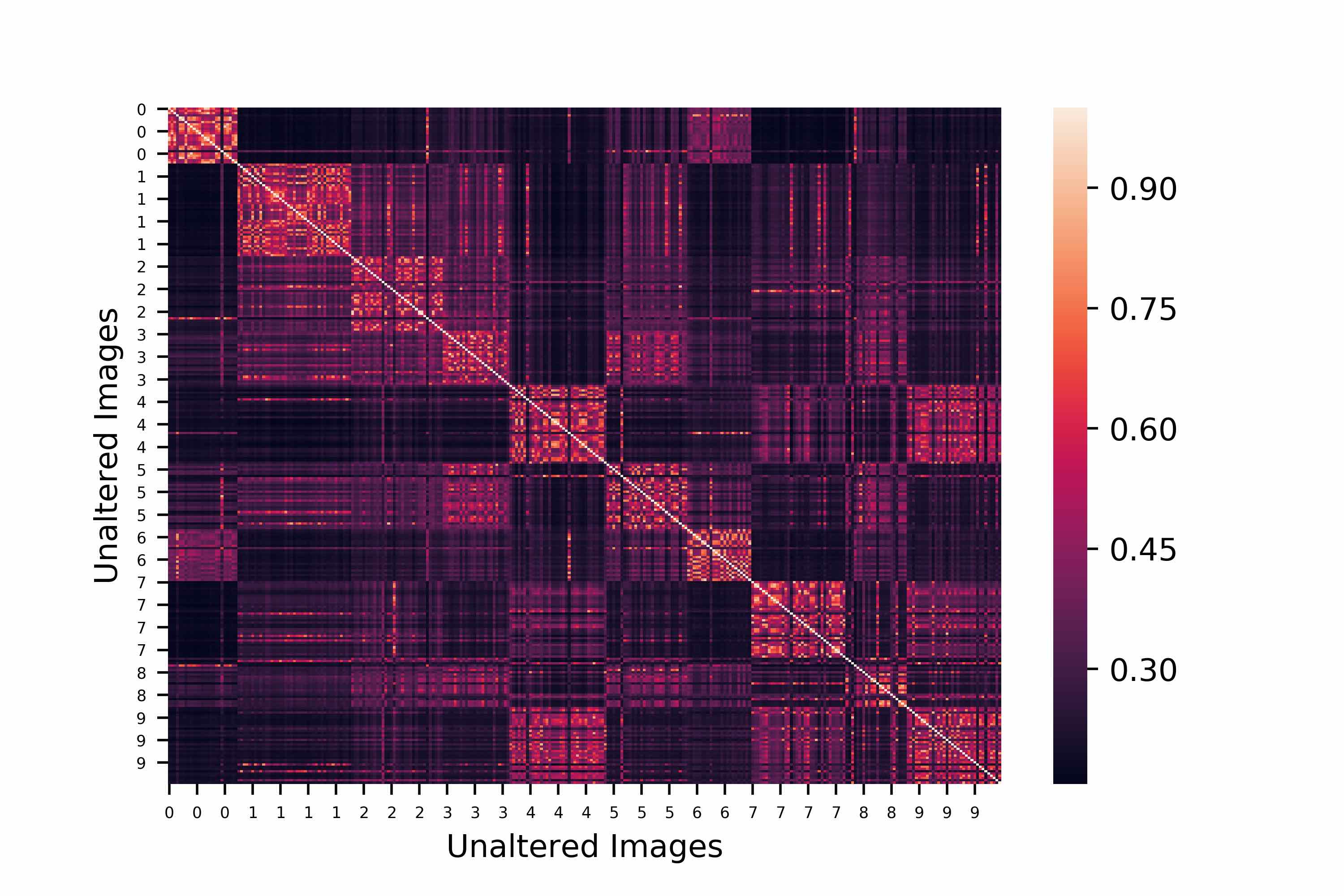}}
        \raisebox{-\height}{\includegraphics[width=0.32\textwidth]{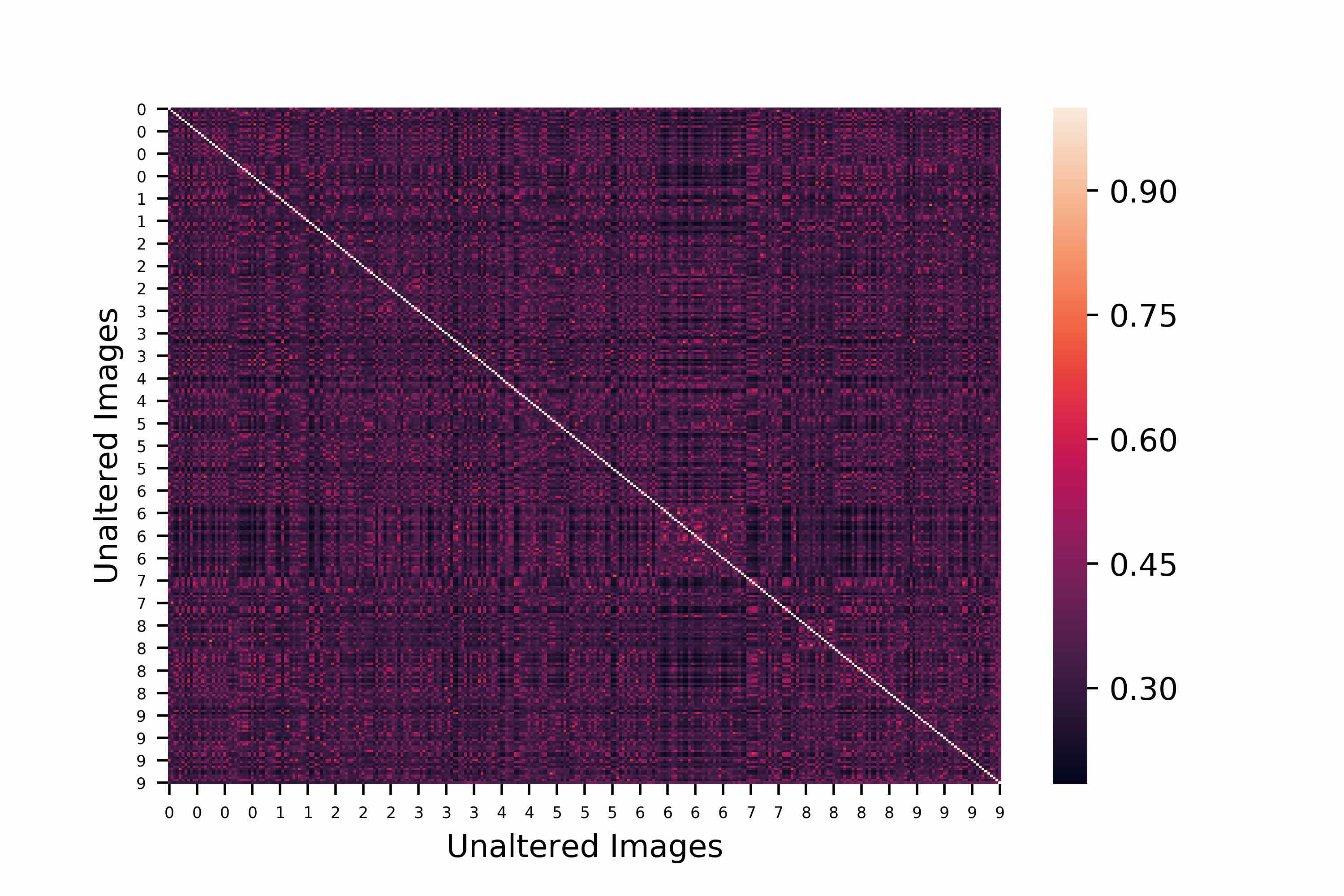}}
        
        \raisebox{-\height}{\includegraphics[width=0.32\textwidth]{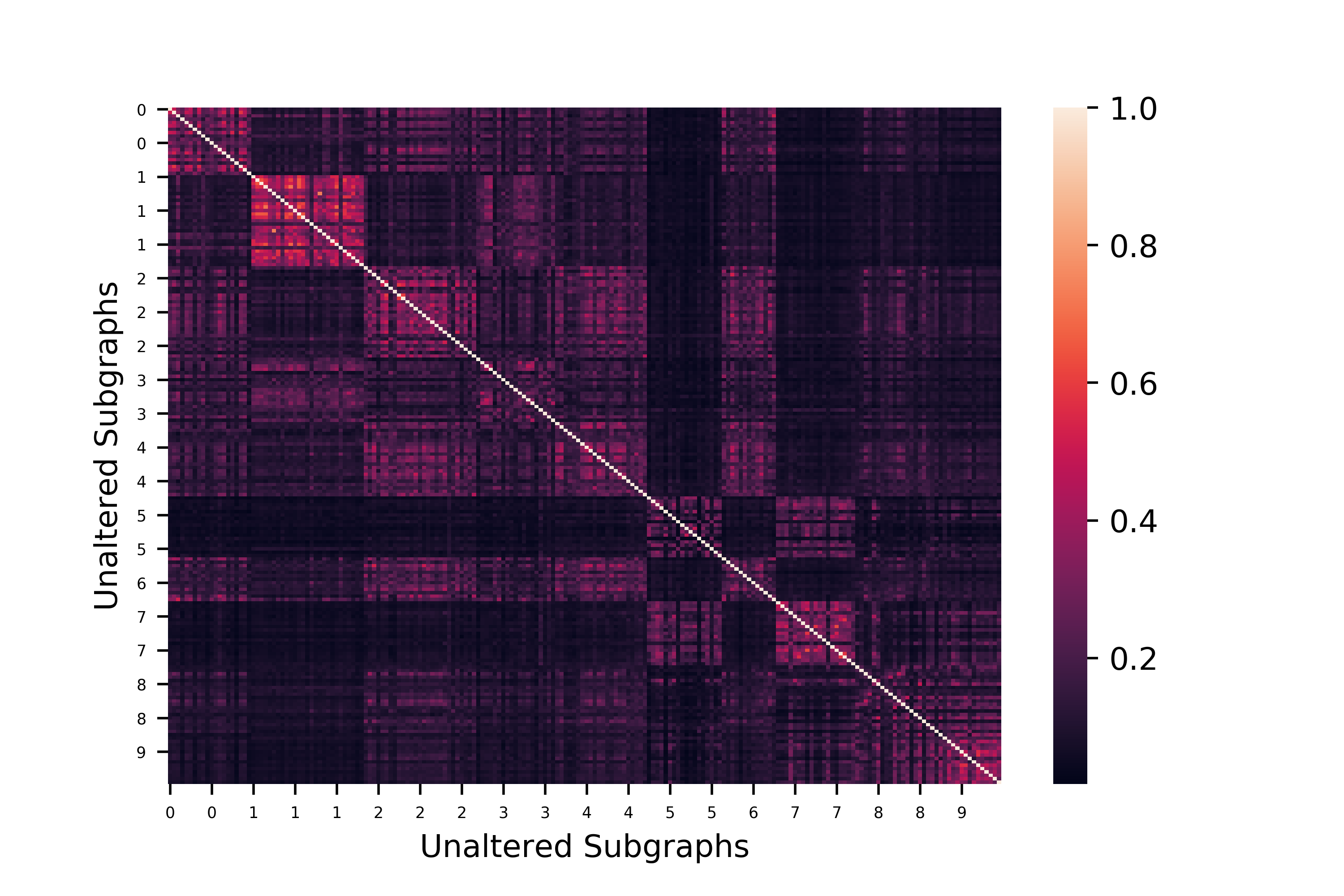}}
        \raisebox{-\height}{\includegraphics[width=0.32\textwidth]{CCFFRelu/nn_ccff_fashion_relu}}
        \raisebox{-\height}{\includegraphics[width=0.32\textwidth, height=3cm]{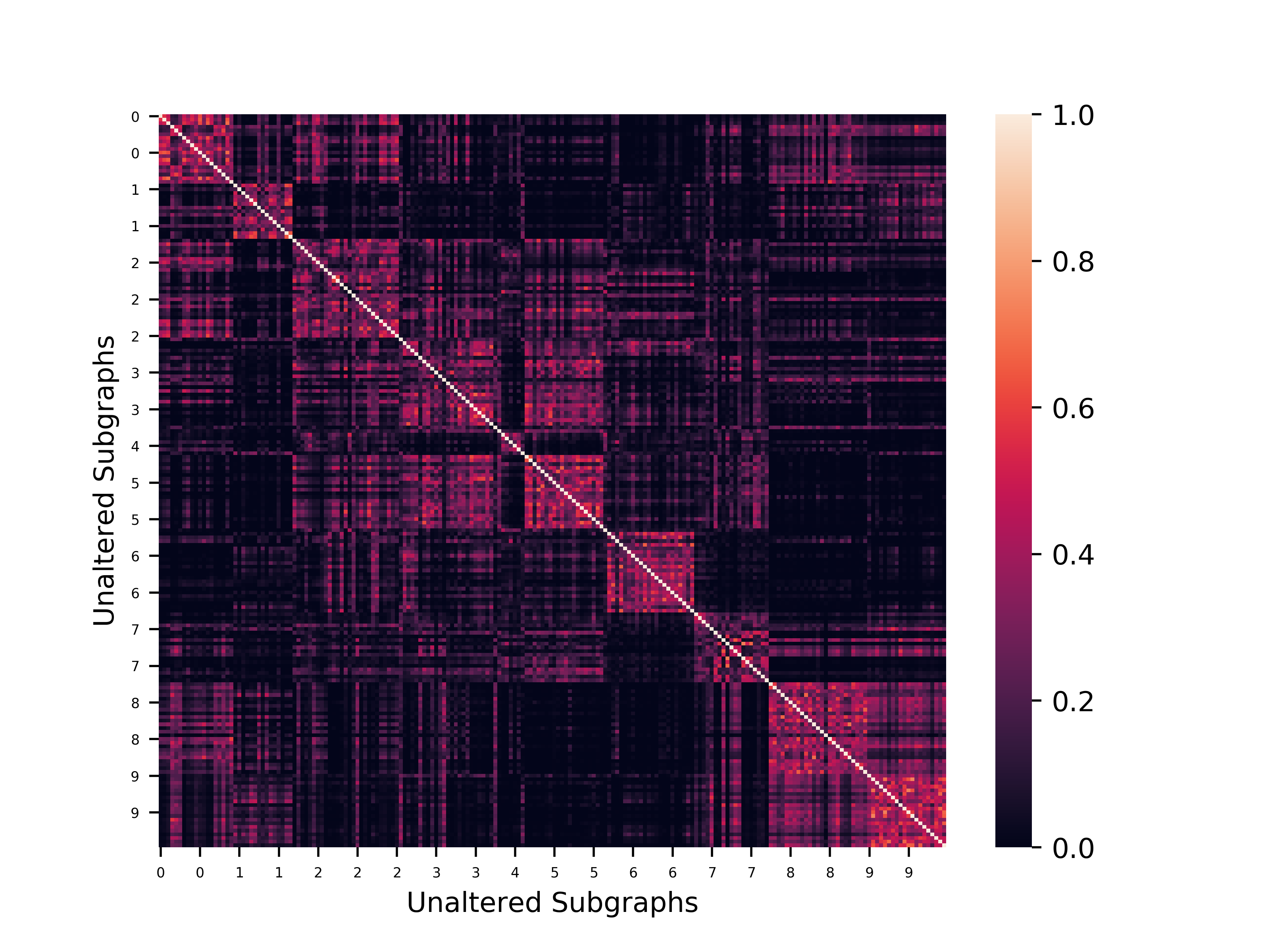}}
\caption{Similarity in input space (top) and persistent subgraph space (bottom) for FashionMNIST (left), MNIST (center), and CIFAR10 (right).}
\vspace{-0.5cm}
\label{fig:neighbors}
\end{figure}

We see from Figure \ref{fig:neighbors} that the subgraph similarity structure largely reflects the similarity structure in image space. Cross-class similarity also shows up within the subgraph similarity structures, implying the existence of representations that are shared across classes. This sharing is desirable, implying the network is learning general representations that can suit the representation of multiple classes before being incorporated into more complex representations in later layers. However, it is unclear whether this representational overlap learned by the networks is optimal. For example, we see that the network representations of the MNIST class 1 are mostly distinct with respect to the representations for all other classes. This could be a sign that the representations learned for this class are too specific to that class and the network's generalization performance would benefit from building this class from representations learned for other classes or vice versa. In general, although we see a structural relationship between similarity in input space and similarity in activation space, the similarities in representation space are not as cohesive within classes as in input space, potentially implying a subtle kind of overfitting within the network where the subgraph representations learned are too specific to regions of input space that could otherwise be spanned by a more general set of representations for a given class. % we see that these are related from figure \ref{fig:di}

\subsection{Adversarial Perturbations}
\begin{figure}[t]
\centering
        \footnotesize{\textbf{CCFF-Relu}} \hspace{0.2\textwidth} \footnotesize{\textbf{CCFF-Sigmoid}} \hspace{0.2\textwidth} \footnotesize{\textbf{AlexNet}}
        
        \raisebox{-\height}{\includegraphics[width=0.32\textwidth]{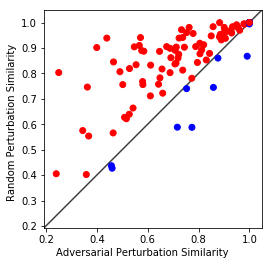}}
        \raisebox{-\height}{\includegraphics[width=0.32\textwidth]{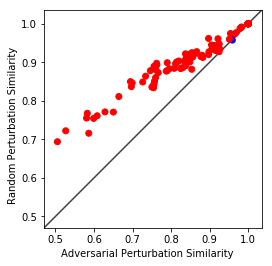}}
        \raisebox{-\height}{\includegraphics[width=0.32\textwidth]{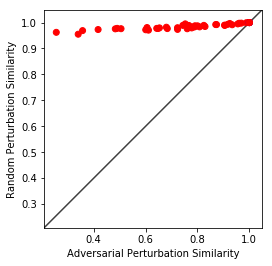}}
        
        \raisebox{-\height}{\includegraphics[width=0.32\textwidth]{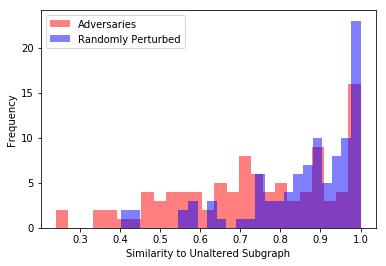}}
        \raisebox{-\height}{\includegraphics[width=0.32\textwidth]{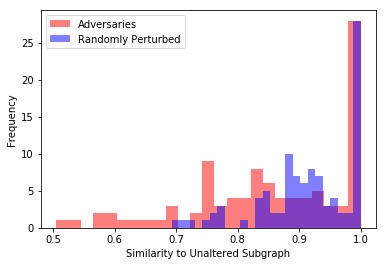}}
        \raisebox{-\height}{\includegraphics[width=0.32\textwidth]{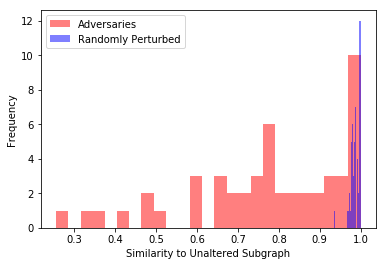}}
        
        \raisebox{-\height}{\includegraphics[width=0.32\textwidth]{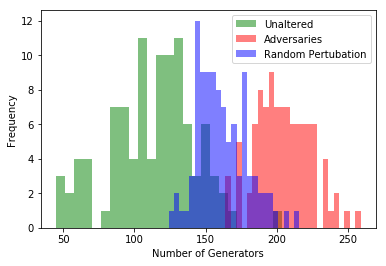}}
        \raisebox{-\height}{\includegraphics[width=0.32\textwidth]{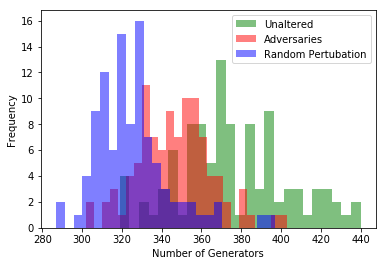}}
        \raisebox{-\height}{\includegraphics[width=0.32\textwidth]{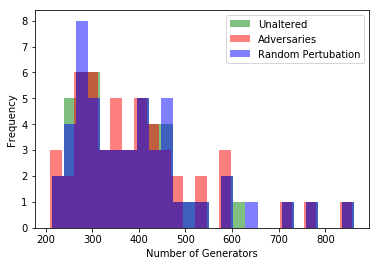}}
        
        \raisebox{-\height}{\includegraphics[width=0.32\textwidth]{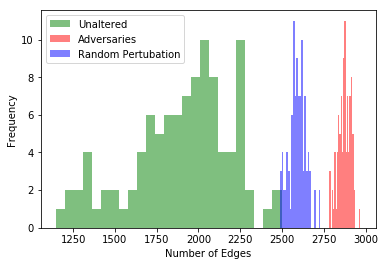}}
        \raisebox{-\height}{\includegraphics[width=0.32\textwidth]{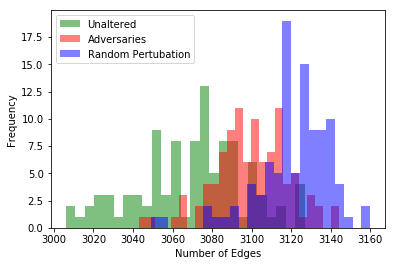}}
        \raisebox{-\height}{\includegraphics[width=0.32\textwidth]{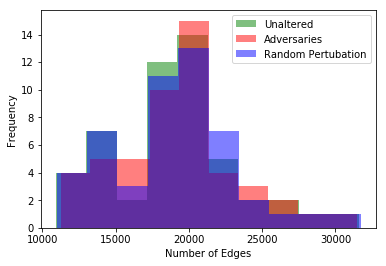}}
\caption{Perturbation effects on subgraph structure for each model. Top Row: Lifetime-weighted similarity to unaltered image for each adversarial example and a random  perturbation of equal magnitude. Second Row: Histogram of top row's similarity information. Third Row: Distribution of the number of generators produced from persistent homology on each type of input. Bottom Row: Distribution of number of edges contained within the generated subgraph structure for each type of input.}
\vspace{-0.5cm}
\label{fig:adversarial_perturbations}
\end{figure}

If a network were creating general representations that properly and robustly partitioned input space, we would expect adversarial perturbations to target the representational structures of the target class, inducing persistent subgraphs associated to the semantics of the target class. However, we find that adversarial examples do not target networks in this way, and that their perturbations are not associated with the representations of any particular class (Supplementary Materials). Instead, adversarial perturbations act in such a way to enact downstream effects in the network \emph{via} the subgraph structure induced by the underlying semantics of the image. In other words, adversarial examples target changes in the input that will flow with the subgraph structure of the input image but cause changes in later layers and, crucially, the output layer. This unique perspective for effects of adversarial examples fits with the previous hypotheses on the subject. In \cite{goodfellow2014explaining}, the authors note that direction is an important determinant of where adversarial examples may be found in image space starting from a given input. From Figure \ref{fig:adversarial_perturbations}, we see that direction is important, in that an arbitrary movement in input space lead to different subgraph structures (in terms of number of edges and generators) as do adversarial movements of equal magnitude. This movement is in the direction of a class boundary, but given that adversarial examples do not activate representational structures of the target class, this implies the decision boundaries of the network are sponge-like, improperly generalizing to all regions slightly off of the image manifold. We also note from Figure \ref{fig:adversarial_perturbations} that the global structure of the persistent subgraphs are more detectable as dataset and network complexity increase.

From the perspective of representations embodied within the subgraph structure, it is unclear how adversarial vulnerability may be remedied under current feedforward architectures. There exist ways to push these representations to more robustly span off-manifold regions of the input space, e.g. adversarial training \cite{zhang2018limitations}, but perfect coverage looks to be infeasible \cite{shafahi2018adversarial}. The experiments above imply that a network properly resistant to adversarial examples would need to be capable of activating representations related to the task-specific semantics of the input while simultaneously suppressing auxiliary activations irrelevant to the task. It appears that feedforward networks are in need of a global regularizer that promotes better invariance to local changes in representational structure.

\section{Conclusion}
The nonlinearities and expansive parameter spaces of deep networks lead to difficulties in the analysis of their learned representations activated by the input. We introduced in this paper a method for computing persistent homology over the graphical activation structure of neural networks and discussed the theoretical benefits of such a topological perspective. We find this graded filtration of subnetworks is effective in capturing local-to-global shape characteristics of the network and can provide insight into some peculiarities of deep network functionality, including the existence of adversarial examples.

\bibliographystyle{plain}
\bibliography{references}

\clearpage

\section{Appendix}

\subsection{Persistent Homology}

Persistent homology provides a method for computing topological features of a space across arbitrarily many resolutions of the space. The topological features that persist across multiple spatial scales may be interpreted as being ``true'' topological features of the space under study. The decomposition, or filtration, of the space corresponds to a choice of metric between points in the space. In the discrete setting, the space is typically represented by a simplicial complex with the filtration describing how the space is constructed based on the chosen metric.

\subsubsection{Simplicial Homology}

Let $\mathcal{K}$ be a simplicial complex. Under $\mathbb{Z}_2$ coefficients, a $p$-chain is a
subset of $p$-simplices of $\mathcal{K}$. The set of $p$-chains, together with addition, forms a free abelian group $C_p$ called the $p$-th chain group of $\mathcal{K}$. The boundary $\partial_p(\sigma)$ of a $k$-simplex $\sigma$ is the set of its $(p-1)$-faces, and that of a $p$-chain is the addition of the boundaries of its simplices. The boundary operator defines a homomorphism $\partial_p : C_p \rightarrow C_{p-1}$.

Define a $p$-cycle to be a $p$-chain with empty boundary and a $p$-boundary to be a $p$-chain
in the image of $\partial_{p+1}$. The collection of $p$-cycles and $p$-boundaries are called the $p$-th cycle group $Z_p = \ker \partial_p$ and the $p$-th boundary group $B_p = \ima \partial_{p+1}$, respectively. Both are subgroups of $C_p$. The \emph{$p$-th homology group} is the quotient group $H_p(\mathcal{K}) = Z_p/B_p$. Elements of $H_p$ are the homology classes $\alpha + B_p = \{\alpha + b \ | \ b \in B_p \}$ for $p$-cycles $\alpha$. We refer to $\alpha$ as the \emph{generating cycle} of the homology class $[\alpha] = \alpha +B_p$. Two $p$-cycles $\alpha$ and $\gamma$ are homologous if $[\alpha] = [\gamma]$, that is, $\alpha+\gamma \in B_p$ is the boundary of some $(p + 1)$-chain.

The rank of $H_p$ is called the $p$-th Betti number of $\mathcal{K}$, denoted by $\beta_p$. A basis of $H_p$ is a minimal set of homology classes that generates $H_p$. A set of $p$-cycles $A = \{\alpha_1, \alpha_2, \dots , \alpha_n \}$ \emph{generates} $H_p$, if the set of generators $\{[\alpha_i]\}$ forms a basis for $H_p$ where $|A| = \beta_p$.

\subsubsection{Persistent Homology}

Let $\mathcal{K}$ be a simplicial complex. A \emph{filtration} is a nested sequence of subcomplexes $\emptyset = \mathcal{K}_0 \subset \mathcal{K}_1 \subset \dots \subset \mathcal{K}_n = \mathcal{K}$. In other words, the filtration is a description of how we want to construct $\mathcal{K}$ by adding arbitrary-sized chunks at a time. With persistent homology, we are interested in the topological evolution of $\mathcal{K}$ as it is deconstructed, which is
expressed by the corresponding sequence of homology groups. Since $\mathcal{K}_i \subset K_{i-1}$, the inclusion of each subcomplex into the larger complex induces a homomorphism between homology groups, $f_p : H_p(\mathcal{K}_{i-1}) \rightarrow H_p(\mathcal{K}_i)$. The nested sequences of complexes shown above thus corresponds to a sequence of homology groups connected by homomorphisms, $0= H_p(\mathcal{K}_0) \rightarrow H_p(\mathcal{K}_1) \rightarrow \dots \rightarrow H_p(\mathcal{K}_n) = H_p(\mathcal{K})$. There exists this sequence of homology groups for each dimension $p$. The filtration defines a partial ordering on the simplices with $\sigma \subset \mathcal{K}_i - \mathcal{K}_{i-1}$ preceding $\tau \subset \mathcal{K}_j - \mathcal{K}_{j-1}$ if $i < j$. This extends to a total ordering for appropriate choice for how simplices are ordered for each $\mathcal{K}_i - \mathcal{K}_{i-1}$. The
rank of $\ima f_p$ is the number of $p$-dimensional homology classes that are born at or before $\mathcal{K}_i$ and are still alive at $\mathcal{K}_j$. We can encode each $h \in \ima f_p$ as a point in the half plane where the $x$-axis encodes the level in the filtration $i$ where $h$ first has a preimage in $H_p(\mathcal{K}_i)$, and the $y$-axis encodes the earliest level in the filtration $j$ where the image of $h$ in $H_p(\mathcal{K}_j)$ is trivial. This half-plane representation is called a \textit{persistence diagram}, denoted $\text{Dgm}(f_p)$.

For two functions $f_p$ and $g_p$, we can compare their persistence diagrams using the \textit{Wasserstein distance} which is defined as the $q$-th root of the
infimum, over all matchings between the points, of the sum of $q$-th powers of the distance between matchings: $$W_q(\text{Dgm}(f_p), \text{Dgm}(g_p) = \inf\limits_\nu \left(\sum\limits_{u \in \text{Dgm}(f_p)} \| u - \nu(u) \|_\infty^q \right)^{\frac{1}{q}}$$ where $q$ is a positive, real number. In the limit for $q$ going to infinity, we get the \textit{bottleneck distance} which is the length of the longest edge in the best matching. Persistence diagrams are stable with respect to the bottleneck distance \cite{cohen2007stability}.

\subsection{Experiments}\label{Experiments}

We implement the ideas provided in the previous sections using three neural network architectures of reasonable size and three datasets. These experiments highlight peculiarities in the representations learned by deep networks and suggest routes for improving their robustness and generalization capabilities. 

\subsubsection{Architectures}

We trained three architectures, two on MNIST and Fashion MNIST, and one on the CIFAR10 dataset. All networks were trained for 20 epochs using stochastic gradient descent with learning rate $0.01$. The first architecture, CCFF-Relu, is a four layer network consisting of two convolutional layers with three 5x5 filters, three 3x3 filters, and ReLU activations followed by two fully-connected layers of size 1452x256 and 256x10 also with ReLU activations. The second architecture is identical to the previous but substitutes sigmoid for ReLU activations. We refer to this network as the CFF-Sigmoid network. The final network is an AlexNet \cite{krizhevsky2014one} variant with zero padding on all kernels and strides appropriately augmented to match this difference. 

\subsubsection{Subgraph Classification}\label{Subgraph Classification}

\begin{figure}[b]
 \centering
  \includegraphics[width=0.47\textwidth]{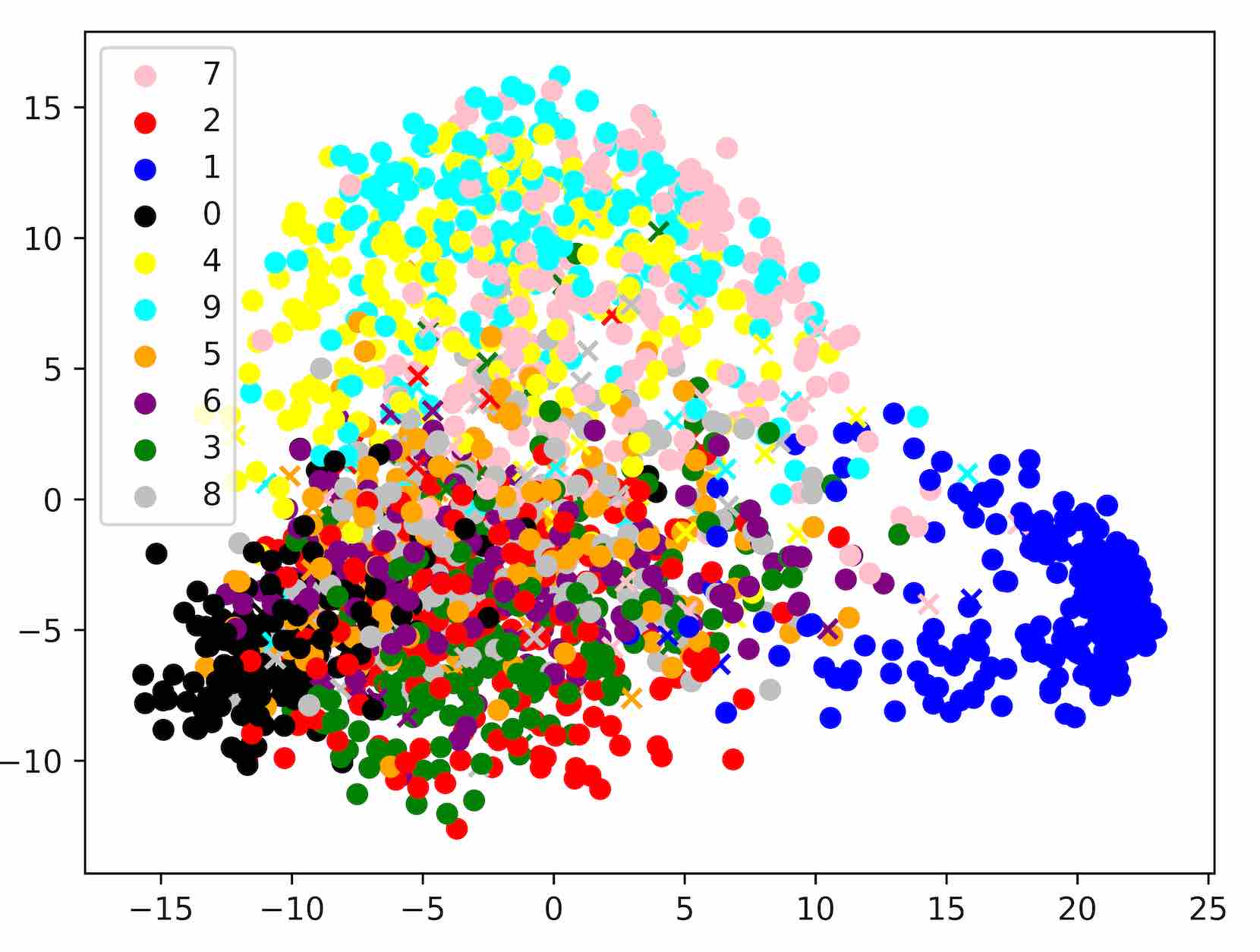}
  \includegraphics[width=0.47\textwidth]{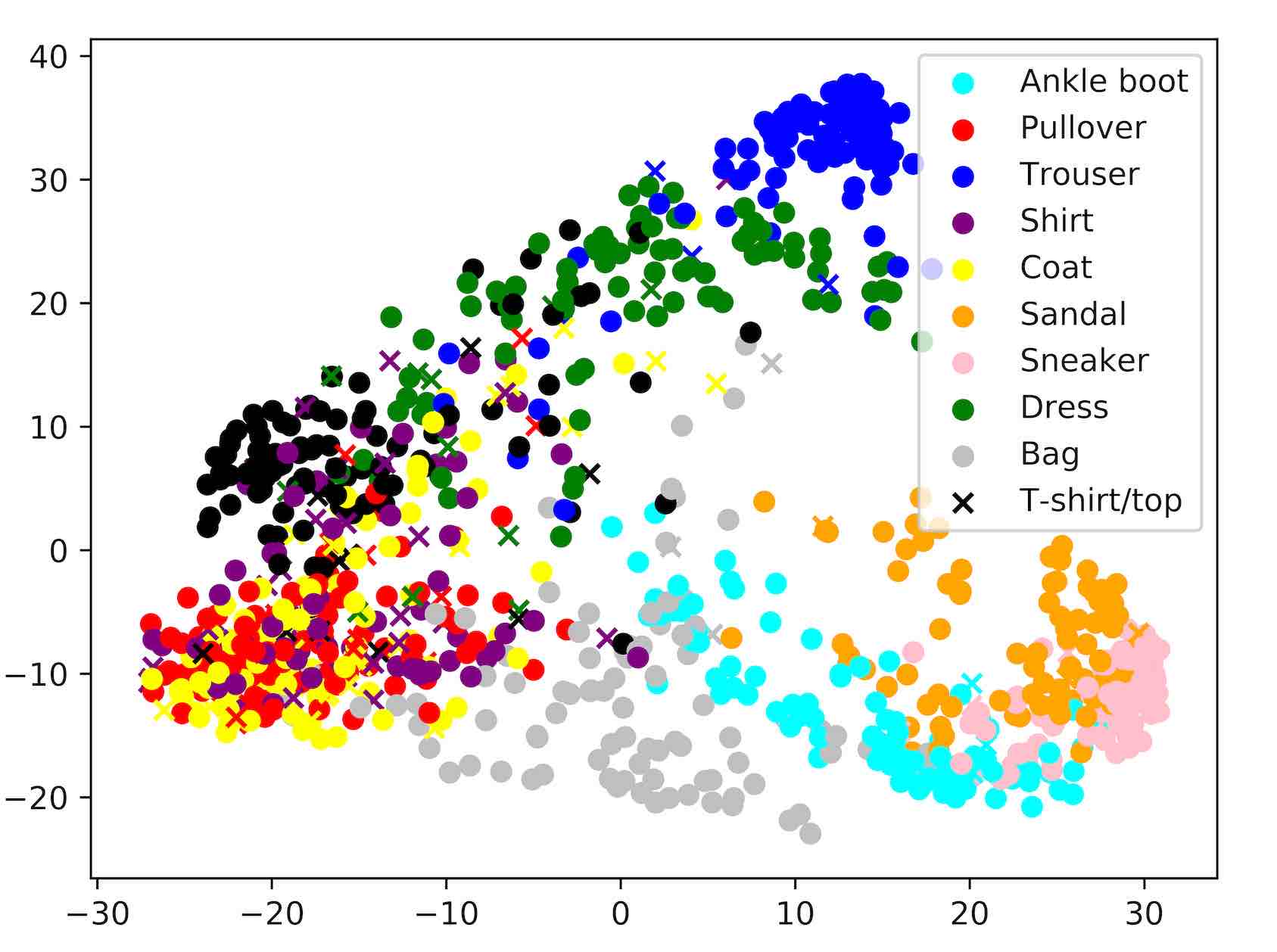}
 \caption{PCA projection of vectorized persistent subgraphs for CCFF-Sigmoid network on MNIST (left) and CCFF-Relu network on Fashion MNIST (Right). Points are colored according to their true class, and misclassified points are plotted as x's. It is clear these subgraphs represent relevant semantic information about the input given the separation of dissimilar objects (0's and 1's) and clustering of like objects (Sandals, Sneakers, Ankle boots).}
 \label{fig:PCA}
 \end{figure}

We perform two classification tasks. For the first, we investigate the extent to which these subgraphs represent class-specific information encoded by the network activation structure. For this, we segment $\mathcal{U}$ into training and test sets $\mathcal{U}_{\text{train}}$ and $\mathcal{U}_{\text{test}}$, and compute the associated vectorization $X_{\text{train}}$. We then vectorize the test set according to the edges corresponding to the dimensions of $X_{\text{train}}$, resulting in matrix $X_{\text{test}}$. For each dataset, architecture, and adversarial generation method, we train an SVM to predict the class of each input represented by $X_{\text{test}}$. The mean accuracy across 10-fold cross-validation is reported in the main paper.

Fix a network architecture, dataset, and adversary generation method. Let $\mathcal{U}$ be the set of unaltered images and $\mathcal{A}$ the set of adversarial images. Also define $U = \{S^\mathcal{I} \ | \ \mathcal{I} \in \mathcal{U}\}$ as the set of persistent subgraphs computed from each unaltered image. Let $\Lambda = \{S^\mathcal{I} \ | \ \mathcal{I} \in \mathcal{A} \}$ be the persistent subgraphs for each adversarial image. Define $U_{\text{train}} \subset U$ as a training subset of unaltered induced subgraphs. Let $N = | U_{\text{train}}|$ be the size of the training set. Each training set image has a corresponding class label representing the true class of the input image. We also create a simple one-hot edge occupancy vectorization for the entire training set $U_{\text{train}}$, where each unique edge in $U_{\text{train}}$ is represented by a dimension in the vectorization, and induced subgraphs $S_{\text{train}}^\mathcal{I} \in U_{\text{train}}$ have value $1$ along a dimension if they contain that edge. Let 
\begin{equation*}
D = \left| \ \bigcup\limits_{E \in U_{\text{train}}} E \ \right|
\end{equation*}
be the size of this vectorization. The result is a matrix $X_{\text{train}} \in \mathbb{Z}_2^{N \times D}$. Example PCA projections of this matrix are shown in Figure \ref{fig:PCA}. 

We train a simple one-versus-one Support Vector Machine (SVM) \cite{cortes1995support} with the lifetime-weighted kernel. Using this classifier, we perform two classification tasks. For the first, we investigate the extent to which these subgraphs represent class-specific information encoded by the network activation structure. For this, we segment $\mathcal{U}$ into training and test sets $\mathcal{U}_{\text{train}}$ and $\mathcal{U}_{\text{test}}$, and compute the associated kernelization $X_{\text{train}}$. We then kernelize the test set according to the edges corresponding to the dimensions of $X_{\text{train}}$, resulting in matrix $X_{\text{test}}$. For each dataset, architecture, and adversarial generation method, we train an SVM to predict the class of each input represented by $X_{\text{test}}$.

For the second classification task, we are interested in the extent to which the network activation structure is able to retain information about adversarial inputs, despite misclassification. We take $\mathcal{U}_{\text{train}} = \mathcal{U}$ and take $\mathcal{A}$ to be the test set. We again kernelize the subgraphs through the training edge set. For each dataset and adversary method, we train an SVM to predict the class of the input based only off of the persistent subgraph information. The accuracy of the original network on $\mathcal{U}$, the accuracy of the SVM in recovering the true class from an adversarial input (Recovery Accuracy) is reported in the main paper. All networks have Network Accuracy of 0\% on their adversary test set $\mathcal{A}$. 

Note that, in all subgraph classification experiments, we do not do hyperparameter tuning of any sort. This simplistic model setup was chosen as a proof of concept in the recognition of input semantics via network subgraph structure. It is expect that much better results could be achieved with a more sophisticated modeling technique and proper parameter tuning.

\subsubsection{Adversarial Perturbations}

\begin{figure}
\centering
        \raisebox{-\height}{\includegraphics[width=0.32\textwidth]{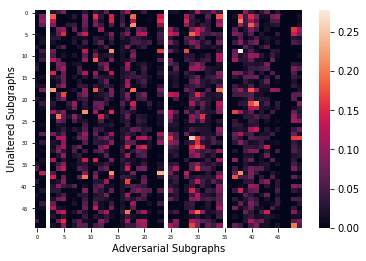}}
        
        \raisebox{-\height}{\includegraphics[width=0.32\textwidth]{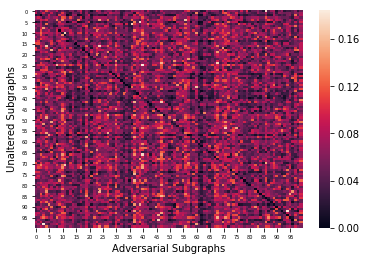}}
        \raisebox{-\height}{\includegraphics[width=0.32\textwidth]{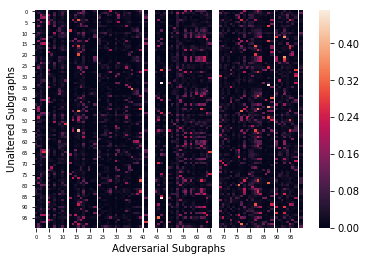}}
        
        \raisebox{-\height}{\includegraphics[width=0.32\textwidth]{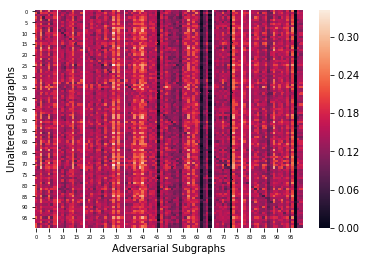}}
        \raisebox{-\height}{\includegraphics[width=0.32\textwidth]{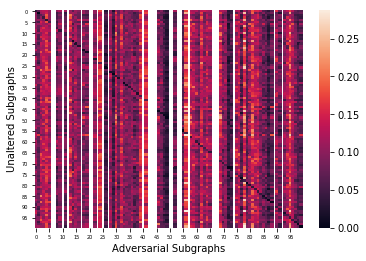}}
\caption{Edge set difference similarity structures for AlexNet on CIFAR10 (top), CCFF-Relu (middle row), and CCFF-Sigmoid (bottom row). The left column is MNIST and the right column is Fashion MNIST.}
\vspace{-0.5cm}
\label{fig:sim_diffs}
\end{figure}

We compute the adversarial perturbations via Euclidean distance in input space and compute the associated random perturbation by adding Gaussian noise to the underlying image until the norm of the noise equals the norm of the adversarial perturbation. We then compute the induced persistent graph structure for each image, its randomly perturbed image, and its adversarial image and display the some of the structural differences in the main figure. We find that the majority of adversarial examples induce more edges within the subgraph structure than their unaltered counterpart (100\% for CCFF-Relu, 89\% for CCFF-ReLU, 80\% for AlexNet). We compute for each network and dataset the set differences between the edge sets of the unaltered subgraph structure and the adversarial subgraph structure, leaving us with the persistent edges that are induced by the adversarial image but not the unaltered image. We then compare this set of edges to the induced subgraphs all other unaltered subgraphs. If the adversarial examples were targeting the semantics of their target classes, we would expect the set difference (the added edges) would be found within the persistent subgraphs induced by the unaltered images of that class. Figure \ref{fig:sim_diffs} shows this similarity structure. The rows of the similarity matrix are ordered by the predicted class while the columns are ordered by target class of the adversary. We would expect block structure to emerge if the added edges were found in the persistent subgraphs of the target class as induced by an unaltered image.

\end{document}